# Estimate the building height at a 10-meter resolution based on Sentinel data

Xin Yan[1*]

[1*]*Department of Geography, University of Georgia, Athens, GA, U.S.A.*

Corresponding author's e-mail: xin.yan@uga.edu

**Abstract**: Building height is an important indicator for scientific research and practical application. However, building height products with a high spatial resolution (10m) are still very scarce. To meet the needs of high-resolution building height estimation models, this study established a set of spatial-spectral-temporal feature databases, combining SAR data provided by Sentinel-1, optical data provided by Sentinel-2, and shape data provided by building footprints. The statistical indicators on the time scale are extracted to form a rich database of 160 features. This study combined with permutation feature importance, Shapley Additive Explanations, and Random Forest variable importance, and the final stable features are obtained through an expert scoring system. This study took 12 large, medium, and small cities in the United States as the training data. It used moving windows to aggregate the pixels to solve the impact of SAR image displacement and building shadows. This study built a building height model based on a random forest model and compared three model ensemble methods of bagging, boosting, and stacking. To evaluate the accuracy of the prediction results, this study collected Lidar data in the test area, and the evaluation results showed that its R-Square reached 0.78, which can prove that the building height can be obtained effectively. The fast production of high-resolution building height data can support large-scale scientific research and application in many fields.

**Keywords**: Remote sensing; building height; Sentinel satellite; random forest model; model ensemble; feature selection

## Introduction

Cities are the centers of human social and economic activity (X. Zhang et al., 2022). To accommodate the increasing population and save land resources, the buildings are inevitably growing on a three-dimensional scale, reflecting the increment in building height. In addition, building height is an important indicator for estimating energy consumption, material stock allocation, greenhouse gas emissions, urban heat island effects, distribution of population, and urban development plan. However, building height products with a high spatial resolution (10m) are still very scarce. Traditional in situ field surveys are an effective way to collect building information. However, it's both time and labor-consuming and hard to update at a large scale frequently. The research in building height estimation has been rapidly advancing in recent years, with new techniques and technologies developed to increase accuracy and efficiency. It is mainly reflected in the combined use of multiple data sources and the improvement of building height estimation algorithms.

The increasing availability of high-resolution satellite images has provided a better source for building height extraction than aerial photographs because they provide several advantages, including cost and accessibility. The satellite images commonly used in this field can be divided into SAR and optical. However, they have different imaging principles and resolutions (spatial, temporal, and spectral), which makes estimating building height more complex. The building height estimation methods vary according to the data, but it has indeed gone through rapid development in recent years. Colin-Koeniguer investigated the use of polarimetry to improve the estimation of the height of buildings in high-resolution synthetic aperture radar (SAR) images (Colin-Koeniguer & Trouvé, 2014). The use of deep learning algorithms (Cao & Huang, 2021), such as convolutional neural networks (CNNs), can be used to estimate building heights from aerial or satellite imagery (Amirkolaee & Arefi, 2019; Mou & Zhu, 2018). Wu et al. proposed an approach to estimate the 2020 building height for China at 10 m spatial resolution based on all-weather earth observations (radar, optical, and night light images) using the Random Forest (RF) model (Wu et al., 2023). Computer vision and machine learning techniques improve understanding of the urban environment and support more accurate building height estimates (Liu et al., 2020). A volumetric shadow analysis (VSA) method was proposed previously for the extraction of 3D building information (height, shape, and footprint location) and for handling occluded building footprints or shadows (Lee & Kim, 2013). However, the spatial resolution of the building height products is coarse (500m or coarser) among the existing models, limiting their broader applicability. On the other hand, the complexity of urban environments, including shadows, occlusions, and overlapping structures, can introduce challenges in estimating building heights accurately. For example, the double bounce (secondary scattering) phenomenon greatly affects the result, and the above studies didn't consider this issue and its solution. All the secondary scattering echo energy will superimpose and converge at the bottom of the dihedral angle formed by the wall and the ground, thus generating a very strong echo signal, which appears as an extremely bright strip in the SAR image. And in many cases, the main contribution to buildings' radar cross section (RCS) comes from various secondary scattering. This phenomenon is especially prevalent in urban areas with densely distributed buildings. Overall, the accuracy of building height estimation algorithms depends on the quality of the input data and the algorithms' robustness.

Existing building height studies and products based on satellite data are limited regarding spatial resolution and coverage. So, the main task of this study is to 1) establish a set of spatial-spectral-temporal feature databases combine the optical, SAR and morphology data; 2) build a robust model to obtain building heights with 10m resolution in the United States; and 3) analyse the building height distribution in Iowa.

**Study area and datasets**

To reduce the overfitting as possible, in model training, this work collected the reference building height ground-truth data from twelve cities: New York City, Boston, Austin, Des Moines, Las Vegas, Richmond, Salt Lake City, Chicago, Oklahoma City, Indianapolis, Baltimore, and Los Angeles. The selection size is 10 km * 10 km. These cities represent megacities to small cities, which can significantly reduce the uncertainty of location selection. The total number of training buildings is 729,109. Based on the model with inclusive training data, the author chose Iowa state as the testing area to evaluate the prediction accuracy. The city boundary of Iowa is downloaded from the Iowa Department of Transportation (https://data.iowadot.gov/datasets/IowaDOT::city-4/about), which is freely shared.

This study used optical and SAR data from Sentinel -1 / 2 at 10 m resolution, downloaded from the Google Earth Engine platform, and all data were pre-processed by the GEE platform. The Sentinel-2 satellite carries a high-resolution multispectral imaging device with 13 bands, including three spatial resolutions of 10m, 20m, and 60m. This study used 4 bands (Band 2, 3, 4, 8) with 10m resolution from 2015 to 2018. As all surface objects have a given and specific spectral profile, and a spectral band alone rarely corresponds with a measurable quantity (Szabo et al., 2016). Band ratios are often used in remote sensing studies, such as Normalized Difference Vegetation Index (NDVI) (Rouse et al., 1974), Normalized Difference Water Index (NDWI) (Gao, 1996), Modified Normalized Difference Water Index (MNDWI) (Xu, 2006), and Land Surface Water Index (LSWI) (Xiao et al., 2004) indexes. I further extracted minimum, maximum, mean, median, and other statistical information in time scale for four bands and their four indices of Sentinel-2. The relevant indexes are defined as follows:

$$MNDWI = \frac{\rho_{Green} - \rho_{SWIR}}{\rho_{Green} + \rho_{SWIR}} \quad (1)$$

$$NDVI = \frac{\rho_{NIR} - \rho_{Red}}{\rho_{NIR} + \rho_{Red}} \quad (2)$$

$$NDWI = \frac{\rho_{Green} - \rho_{NIR}}{\rho_{Green} + \rho_{NIR}} \quad (3)$$

$$LSWI = \frac{\rho_{NIR} - \rho_{SWIR}}{\rho_{NIR} + \rho_{SWIR}} \quad (4)$$

The Sentinel-1 satellite provides data from a dual-polarization C-band Synthetic Aperture Radar (SAR) instrument at 5.405GHz, which can penetrate clouds and fog and provide continuous images throughout the day and in all weather. Four imaging modes provide different resolutions and coverage; the Interferometric Wide Swath mode, among them, provides a large swath width of 250 km at a ground resolution of 5m×20m, with enhanced image performance compared to the conventional ScanSAR mode. This study adopts the VV (vertical transmit/vertical receive) and VH (vertical transmit/horizontal receive) polarization types. The VVH index based on the methods of the literature is also considered (X. Li et al., 2020).

$$VVH = VV * \gamma^{VH} \quad (5)$$

The reference Light Detection and Ranging (LiDAR) and Digital Elevation Model (DEM) data are included to get the reference height of the buildings, which can be downloaded from USGS (https://apps.nationalmap.gov/downloader/). Lidar point cloud data are the foundational data for 3DEP in the conterminous U.S. and contain the original three-dimensional information from which the DEM products are derived. Most of the data collected in 2014 and later meet 3DEP specifications for quality level 2 nominal pulse spacing and vertical accuracy.

The building footprint data come from the Microsoft Maps (https://github.com/microsoft/USBuildingFootprints), which contain 129,591,852 computer-generated building footprints derived using computer vision algorithms on satellite imagery. The pixel precision of this product reaches 94%. After collecting all of the building footprint data in the US, this study calculated their minimum bounding geometry in ArcGIS pro 3.0.0, including width, length, and orientation. Besides that, this study also calculated the near distance between buildings, which can show the building density. Figure 1 shows the vector building footprint distribution in Iowa.

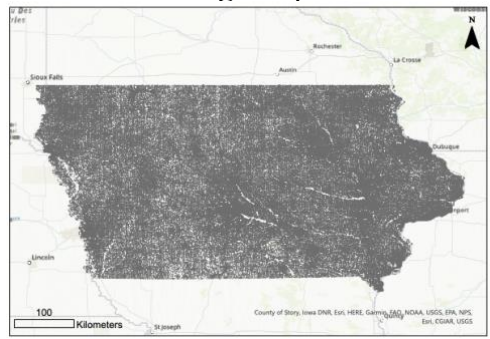

Figure 1. Building footprints in Iowa.

**Methods**

In this study, I developed a method to estimate building height from the Sentinel-1/2 and building footprint data and then calculated the building height of major states in the US. This model mainly included feature selection, construction, and ensemble parts (Fig. 2). The achievement of the final model was built up at each step, contributing to its overall effectiveness and accuracy. In the training part, the building height estimation model is object-based, and all dependent and independent variables are the median values within the range of the building footprint area. However, the resulting building height is a pixel-based median value within the 10 m grid. The object-based model can help improve the training efficiency and reduce the effect of noise, and the pixel-based result can save space for storing the data, which is beneficial for large-scale applications.

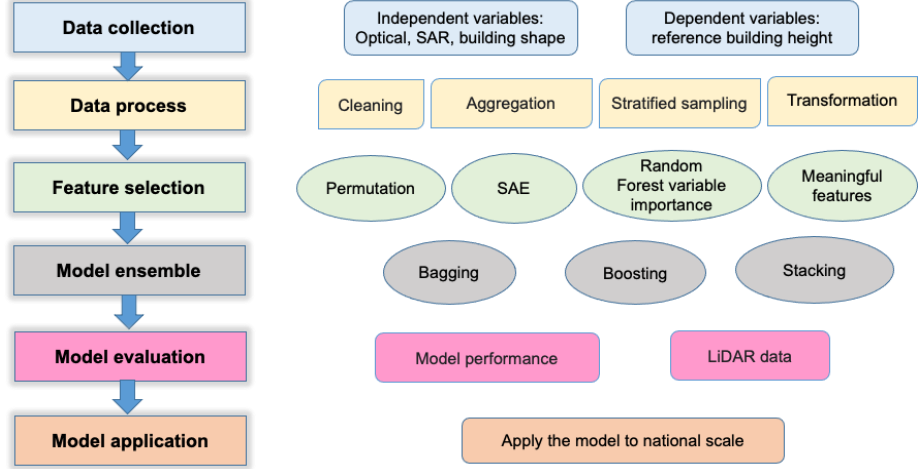

Figure 2. The framework of this work.

*Feature selection*

As a data pre-processing strategy, feature selection has proven effective and efficient in preparing data (especially high-dimensional data) for various data-mining and machine-learning problems (J. Li et al., 2017). It usually can lead to better learning performance, i.e., higher learning accuracy, lower computational cost, and better model interpretability (Miao & Niu, 2016).

*Random forest variable importance ranking*

The random forest can not only realize the classification of remote sensing images but also play an important role in feature selection and dimensionality reduction (Y. Zhang et al., 2019). Samples are usually divided into training and verification samples. The verification samples are called Out-Of-Bag (Out-Of-Bag) because they do not participate in the classification. The Random Forest feature variable importance (VI) is calculated using the Out-Of-Bag-Error (OOBE) generated by the out-of-bag data. The formula is as follows:

$$VI(M_A) = \frac{1}{N}\sum_{t=1}^{N}\left(B_{n_t}^{M_A} - B_{O_t}^{M_A}\right) \quad (6)$$

Where VI represents the importance of feature variables, $M_A$ is a feature, N is the number of decision trees, an $B_{O_t}^{M_A}$ and $B_{n_t}^{M_A}$ are the OOBE of the t-th decision tree when the arbitrary feature values $M_A$ with and without adding noise interference. If noise is randomly added to a certain feature $M_A$, the accuracy of the out-of-bag data is greatly reduced, which means that feature $M_A$ has a great influence on the classification result, and it shows that its importance is relatively high. Feature selection according to their importance ranking is also a commonly used method in feature extraction (Genuer et al., 2010).

*SHapley Additive exPlanations*

SHapley Additive exPlanations, more commonly known as SHAP, are used to explain the output of Machine Learning models. It is based on Shapley values, which use game theory to assign credit for a model's prediction to each feature or feature value. The SHapley value is the average marginal contribution of a feature value across all the possible combinations of features. A key part of this is the difference between the model's prediction with feature i, and the model's prediction without feature i.

$$\emptyset_i = \sum_{S \subseteq M \setminus i} \frac{|S|!(|M|-|S|-1)!}{|M|!} [f(S \cup i) - f(S)] \qquad (7)$$

S refers to a subset of features that doesn't include the feature for which we're calculating $\emptyset_i$, $S \cup i$ is the subset that includes features in S plus feature i, $S \subseteq M \setminus i$ in the $\sum$ symbol is saying, all sets S that are subsets of the full set of features M, excluding feature i.

*Permutation feature importance*

Permutation feature importance directly measures feature importance by observing how random re-shuffling (thus preserving the distribution of the variable) of each predictor influences model performance. A feature is "important" if shuffling its values increases the model error because, in this case, the model relied on the feature for the prediction. A feature is "unimportant" if shuffling its values leaves the model error unchanged because, in this case, the model ignored the feature for the prediction. The following equations can conclude the process:

$$e_{orig} = L\left(y, \hat{f}(X)\right) \qquad (8)$$

$$e_{perm} = L\left(Y, \hat{f}(X_{perm})\right) \qquad (9)$$

$$FI_j = e_{perm} - e_{orig} \qquad (10)$$

$\hat{f}(X)$ is the trained model, X is the feature matrix, y refers the target vector, and the $L\left(y, \hat{f}(X)\right)$ is the measured error. First, we need to estimate the original model error $e_{orig}$ (e.g. mean squared error) based on equation 7; for each feature $j \in \{1, ..., p\}$, degenerate feature matrix $X_{perm}$ by permuting feature j in the data X. This breaks the association between feature j and true outcome y, and then estimate the error $e_{perm}$ based on the predictions of the permuted data (equation 8). Next step is calculating permutation feature importance as a difference based on equation 9. Finally, sort features by descending FI.

*Model ensemble*

Ensemble learning is a general meta-approach to machine learning that combines the predictions from multiple models to achieve better predictive performance. In this study, I compared three main classes of ensemble learning methods (bootstrap aggregation, stacked generalization, and boosting). Bootstrap aggregation, also known as bagging, is a machine-learning ensemble technique that combines the predictions of multiple models to produce a more accurate and stable prediction. The basic idea behind bagging is to create multiple copies of the original dataset by random sampling with replacement. Each copy is then used to train a separate model on the data. Stacked generalization, also known as stacking, is a machine learning ensemble technique that hierarchically combines multiple models to improve the overall prediction accuracy. In stacking, a set of base models is trained on the original data, and their predictions are then used as input features for a higher-level model called the meta-model. Boosting is a machine learning ensemble technique that combines the predictions of multiple weak models to create a strong model that can make more accurate predictions. In boosting, a series of base models, called weak learners, are trained sequentially, with each subsequent model learning from the errors of its predecessor. The outputs are a weighted average of the predictions.

*Model Optimization*

Figure 3 illustrate the causes and manifestations of the double bounce phenomenon. Taking SAR images as an example, when the signal from the satellite hits the target building, it will be reflected in the surroundings due to the influence of the shape and material of the target building, causing the signal strength of the surrounding objects to be higher than its signal strength. In addition, in the optical image, since the imaging time of each image is 10 am, the shadow of the building appears on the other side of the building, which makes the radiation intensity of the image different from that of the ground object itself. This phenomenon has little impact on low- and medium-resolution products, but it becomes one of the non-negligible problems in high-resolution building height prediction.

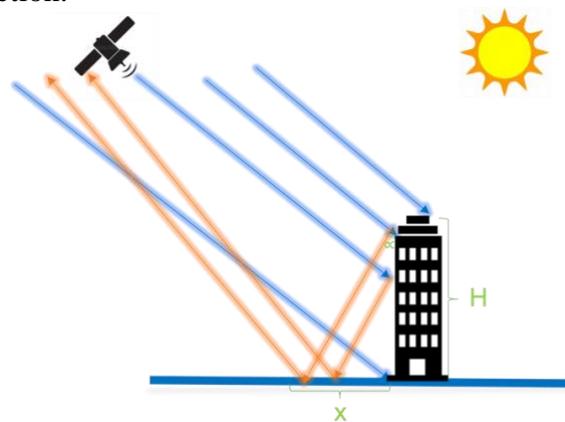

Figure 3. Satellite signal double bounce schematic diagram (H indicates building height, and x indicates the shadow length).

In order to solve this problem, this work built a 50m buffer zone around each building and then collected the median reference height of each zone as the training data. For the final prediction, this work built a 50m*50m moving window for each pixel and calculated the median value of this window as the new pixel value. This study performed a sensitivity analysis of the derived buffer distance and moving window size. The 50m was chosen from 10m, 30m, 50m, 80m, and 100m.

In addition, this study selected the pixels from the 1% - 99% threshold in pixel statistics to remove the noise effect. The height values were also logarithmically transformed, often used to reduce the skewness of a measurement variable. The converted reference height value ranges from 0-6.7, based on which this study divided it into steps of 0.01 and calculated the median within each 0.01 interval. In this way, the representative values of different features in each height interval can be obtained, effectively removing the errors caused by the data.

**Results**

*Preliminary model selection and comparison*

For the model comparison, this study chose 9 machine learning models as a preliminary comparison to run the same 15,000 training and testing samples. The models are Recursive Partitioning and Regression Trees (RPART), Linear models (LM), Generalized Linear Models (GLM), Generalized Linear Models using the Lasso and Elastic-Net penalties (GLMNET), Random Forest regression (RF), K-Nearest Neighbors (KNN), Support Vector Machine with Radial Basis Function (SVM-RBF), Gaussian Process Regression (GPR), Generalized Additive Models (GAM) and Gradient Boosting (GAMBoost). The results show in Table 1. This study chose the first four machine learning methods (GLM, RF, SVM-RBF, and GAMBoost) to do the model ensemble test. After a preliminary model comparison, the random forest model was selected as the training model in our study.

Table 1. The accuracy of different models

| Model | min | 1$^{st}$ Qu | median | mean | 3$^{rd}$ Qu | max |
|---|---|---|---|---|---|---|
| RPART | 0.42 | 0.45 | 0.51 | 0.50 | 0.53 | 0.60 |
| GLM | 0.01 | 0.61 | 0.62 | 0.53 | 0.64 | 0.68 |
| GLMNET | 0.60 | 0.62 | 0.64 | 0.64 | 0.65 | 0.69 |
| LM | 0.01 | 0.61 | 0.62 | 0.53 | 0.64 | 0.68 |
| RF | 0.68 | 0.72 | 0.74 | 0.74 | 0.76 | 0.80 |
| KNN | 0.45 | 0.50 | 0.53 | 0.53 | 0.55 | 0.64 |
| SVM-RBF | 0.68 | 0.71 | 0.73 | 0.73 | 0.75 | 0.78 |
| GPR | 0.60 | 0.61 | 0.63 | 0.63 | 0.65 | 0.69 |
| GAMBoost | 0.65 | 0.66 | 0.68 | 0.68 | 0.69 | 0.72 |

*Feature Database*

To meet the needs of building height estimation models with a resolution of 10 meters, this study has established a set of spatial-spectral-temporal feature databases, combining SAR data provided by Sentinel-1, optical data provided by Sentinel-2, and shape data provided by building footprints. The average, median, standard deviation, percentage, and other statistical indicators on the time scale are extracted to form a rich database of 160 features. At the same time, combined with permutation feature importance, Shapley Additive Explanations, Random Forest variable importance, and manual selection methods, the final stable features are obtained through repeated experiments and an expert scoring system for model training and application. Table 2 shows the selected 13 features.

Table 2. Feature selection results

| Random Forest | Permutation | SHapley Additive exPlanations | Final choice |
|---|---|---|---|
| LSWI_skew | MBG_Width | MBG_Width | LSWI_skew |
| MBG_Width | MBG_Length | NDVI_skew | MBG_Width |
| MBG_Length | NDVI_skew | MBG_Length | MBG_Length |
| LSWI_kurtosis | LSWI_skew | NDWI_skew | LSWI_kurtosis |
| NDVI_skew | MNDWI_skew | NDBI_p100 | NDVI_skew |
| NDWI_skew | VVH_skew | LSWI_p0 | NDWI_skew |
| NDVI_interquatile_range | LSWI_kurtosis | B8_p0 | B3_p10 |
| B4_p10 | NDWI_skew | B6_mean | NDVI_interquatile_range |
| MNDWI_kurtosis | NDBI_p100 | VH_p100 | B4_p10 |
| B3_p10 | VV_p100 | NDVI_p100 | VV_mean |
| VVH_stdDev | B5_p10 | LSWI_kurtosis | VH_mean |
| B5_p10 | NDVI_p100 | B3_p10 | VVH_stdDev |
| NDBI_skew | B3_p10 | LSWI_skew | VVH_mean |

*Model Performance*

In this study, the model performance includes two parts, the training performance of this model and the model's prediction performance. I evaluated the performance of building height models using all reference building height from LiDAR in twelve US cities, and compared the estimated building height. To make sure the accuracy of reference data, I also collected the building height data published from government officially, such as the Des Moines (https://www.dsm.city/city_of_des_moines_gis_data/).

For the training performance of the model, there are 471 processed samples involved in this model because I divided the original samples into 0.01 interval steps based on the log transformation of building heights, which also is designed to consider the training efficiency. The results are shown in Figure 4. In this model, 98.72% of variables can be explained, and the Mean of squared residuals is 0.049. Our model has been proven to have the strong power to predict the building height in 10m resolution.

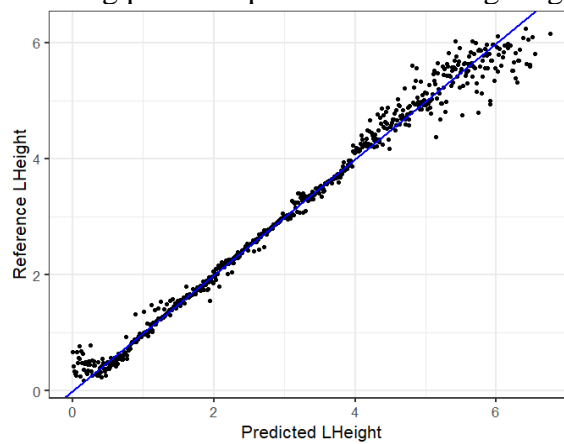

Figure 4. The accuracy of training data.

This study randomly selected 12,000 samples from raw data and predicted their building heights. Figure 5 shows the predicted and reference log transformation of building height. The $R^2$ could reach 0.782, which is an acceptable accuracy compared with the most up-to-date building height products, whose accuracy is usually around 0.6. Like the figure of the training accuracy, the model did well when the building height was between the middle range (LHeight is between 1 and 4) but did worse when the building height was tall.

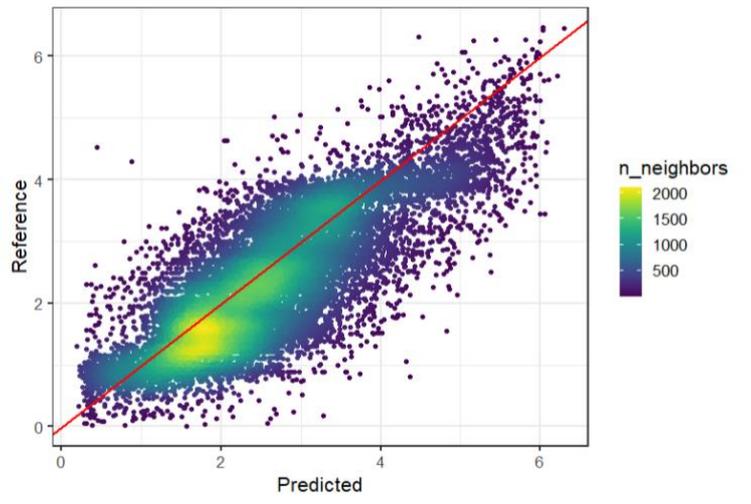

Figure 5. The accuracy of testing data.

*Building Height Distribution in Iowa*

In Iowa, there are a total of 2,074,904 buildings, which cover 2,855,860,281 pixels at 10-m resolution satellite images. The predicted results show that the minimum building height is 1.23m while the highest is 539.68m. The mean value of building height in Iowa is 5.24m. Figure 6 shows the overall building distribution in Iowa and a detailed example in Des Moines.

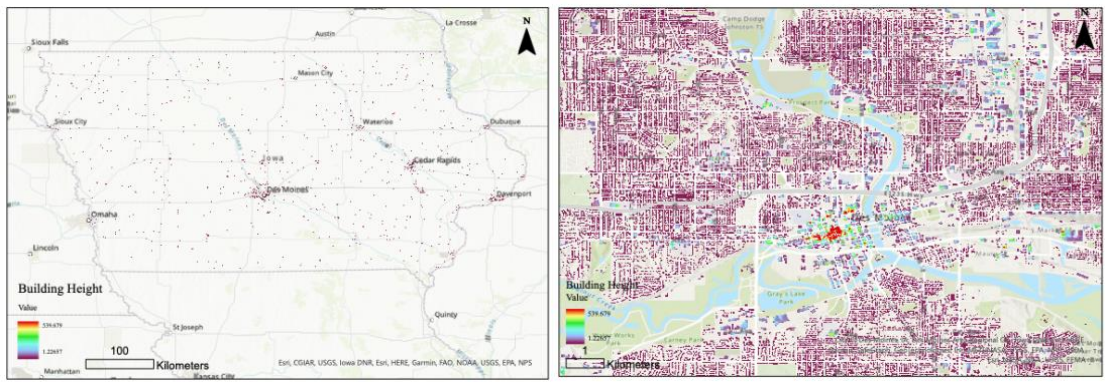

Figure 6. Predicted building height in Iowa (left) and Des Moines (right).

There are 99 counties in Iowa (Figure 7). Geographically, Southeast, Central, and Northwest Iowa have higher average building heights than the state average, while the Southwest has generally shorter building heights. At the county level, four cities have a maximum building height greater than 200: Pottawattamie County, Muscatine County, Wapello County, and Polk county. And Scott County, Polk County, Sioux County and Black Hawk County have the highest mean building height.

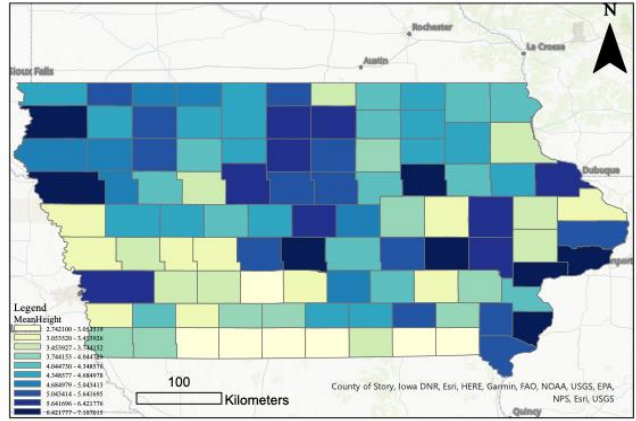

Figure 7. The mean building height distribution in Iowa at the county level.

The counties with a land area greater than 2000 square kilometers, sorted by area, are Woodbury County, Kossuth County, Plymouth County, Pottawattamie County, and Clayton County. But in order of building footprint area, the counties are Polk County, Linn County, Scott County, Black Hawk County, Johnson County, Dubuque County, Pottawattamie County, and Woodbury County. The level of construction development in each county is different. Still, in order of the proportion of construction to land area, the top five counties are Polk County, Scott County, Linn County, Black Hawk County, Johnson County, and Dubuque County.

**Discussion**

Overfitting is a fundamental issue in supervised machine learning, which prevents us from perfectly generalizing the models to fit observed data on training data, as well as unseen data on the testing set. Because of the presence of noise, the limited size of training set, and the complexity of classifiers, overfitting happens (Ying, 2019). We can see from Figures 4 and 5 that the variables on the two sides (low and high height) have higher inaccuracy, which has many possible reasons. The first reason is that the reference height may be inaccurate, especially for high buildings; for example, many buildings counted the height of the antenna in its overall height. The second reason is the double bounce effects in the image. A moving window of 50m*50m will also make smaller buildings over-consider the influence of surrounding pixels. For example, the moving windows make the surrounding non-building pixels be considered to re-calculate the new pixels, thereby reducing or enhancing its original signal intensity, and its influence is mainly based on the surrounding condition of the building. The third reason is that the spatial resolution cannot cover the size of some buildings, resulting in the signal not representing the reality of the building height; that is to say, the height of many buildings does not reach 10m. The fourth reason is related to the building materials; the characteristics of different materials affect the signal strength of SAR images, especially for glass and concrete. The last reason is the shape of the roof; rough surface scattering, such as that caused by a garden or swimming pool, is most sensitive to VV scattering. In addition, the urban pattern and influencing factors of various cities are different, resulting in the model trained in one city being unable to be well applied to other cities. Although this study hopes to weaken this phenomenon by using training data from 12 cities across the country, a model cannot fit all cities all the time. For example, a city in New York versus a Midwestern US city will have different test accuracy.

Considering that I hope to apply this model nationwide or even globally, the calculation of the building shape and the complexity of the model pose a test. Therefore, this article only considers commonly used machine learning models and does not consider deep learning algorithms, although deep learning has demonstrated good application capabilities in various fields. The author will test several deep learning algorithms to comprehensively consider the accuracy and operating efficiency of the model.

Therefore, there are associated uncertainties in the derived building height data from the definition of building height, backscatter coefficients from the Sentinel-1 data, and the proposed building height model. Author will consider different models based on urban density in the following research.

**Conclusion**

To meet the needs of building height estimation models with a resolution of 10 meters, this study has established a set of spatial-spectral-temporal feature databases, combining SAR data provided by Sentinel-1, optical data provided by Sentinel-2, and shape data provided by building footprints. The feature database has a total of 160 features, including the average, median, standard deviation, percentage, and other statistical indicators on the time scale. To get robust model results, this study combined three feature selection methods, which are permutation feature importance, Shapley Additive Explanations, and Random Forest variable importance methods. This study built a building height model using a random forest machine learning model based on the bagging ensemble method. The innovation of the article is mainly to use a building-based rather than a pixel-based method and to use buffering to solve the difficulty of multiple scattering of SAR data. The proposed model has relatively good accuracy and could be used for large-scale applications for fast production of high-resolution building heights. The article uses Iowa as an example to predict building height. High-resolution building height data can support scientific research and production applications in many fields.

**Acknowledgments**

The author would like to thank the reviewers and editors who provided valuable comments and suggestions for this article and the editorial board's support. I also would like to thank Iowa State University and the University of Georgia for providing me with excellent working conditions and the faculties and graduate students in these two universities for their valuable suggestions for my research.